# Pedestrian Crossing Intention Prediction Using Multimodal Fusion Network

Yuanzhe Li and Steffen Müller

Chair of Automotive Engineering, Technische Universität Berlin,
Gustav-Meyer-Allee 25, 13355 Berlin, Germany
yuanzhe.li@campus.tu-berlin.de

**Abstract.** Pedestrian crossing intention prediction is essential for the deployment of autonomous vehicles (AVs) in urban environments. Ideal prediction provides AVs with critical environmental cues, thereby reducing the risk of pedestrian-related collisions. However, the prediction task is challenging due to the diverse nature of pedestrian behavior and its dependence on multiple contextual factors. This paper proposes a multimodal fusion network that leverages seven modality features from both visual and motion branches, aiming to effectively extract and integrate complementary cues across different modalities. Specifically, motion and visual features are extracted from the raw inputs using multiple Transformer-based extraction modules. Depth-guided attention module leverages depth information to guide attention towards salient regions in another modality through comprehensive spatial feature interactions. To account for the varying importance of different modalities and frames, modality attention and temporal attention are designed to selectively emphasize informative modalities and effectively capture temporal dependencies. Extensive experiments on the JAAD dataset validate the effectiveness of the proposed network, achieving superior performance compared to the baseline methods.

## 1 Introduction

With the rapid development of artificial intelligence and sensor technologies, autonomous vehicles (AVs) are being increasingly deployed in urban environments. Pedestrian crossing intention prediction plays a vital role in enhancing the safety of AVs by extending their effective response window [17]. Despite its importance, the prediction task is challenged by behavioral diversity of pedestrians and environmental factors, such as road structure, traffic conditions, and interactions with nearby participants [2].

Pedestrian crossing intention prediction has been extensively studied over the past several years. In [3] convolutional neural network (CNN) is used to extract features from a single image for intention prediction. However, it overlooks the complementary properties of multimodal information, which plays a critical role in the intention prediction task. Recent studies leverage multimodal data to address this task, intro-ducing additional visual modalities (e.g., semantic maps [5], optical flow [6]) and motion modalities (e.g., bounding boxes, vehicle speed [7]) to capture pedestrian-specific and global contextual information. Most studies design separate branches for each modality to extract features from the whole observation sequence, followed by modality feature fusion for intention prediction. For instance, in [8], MCIP is proposed, where CNNs are employed to extract visual features from local image and global semantic map, which are subsequently pooled and individually encoded together with data from other motion modalities by separate recurrent neural networks (RNNs) for temporal fusion. Finally, a modality attention mechanism is applied to integrate the temporally fused features from the visual and motion branches.

In this paper, we propose P-MFNet, a multimodal fusion network designed for pedestrian crossing intention prediction. Our main contributions are: (1) Multiple Transformer-based feature extraction modules are developed to effectively extract visual features and motion features. (2) Depth-guided attention mechanism is introduced for spatial feature fusion, using the depth map to guide attention toward salient regions in the local RGB image and global semantic map. (3) Modality attention module is designed to model each modality's importance, while temporal attention module is introduced to capture temporal dependencies. (4) Extensive experiments on the JAAD dataset are carried out to demonstrate the performance of P-MFNet.

## 2 Problem Formulation

The task is formulated as binary classification to predict whether a pedestrian will cross within $f$+$n$ frames ($n\in\{30, 60\}$, i.e., 1–2s ahead). The prediction is based on analyzing a sequence of video frames observed over the past 0.5 seconds. P-MFNet integrates seven visual and motion modalities, including: (1) Pedestrian's pose keypoints $P=\{p^{f-N+1}, p^{f-N+2},\ldots, p^{f}\}$: 2D coordinates of 18 joints extracted using a pretrained OpenPose model [9], providing posture cues. (2) Pedestrian's bounding box $B=\{b^{f-N+1}, b^{f-N+2},\ldots, b^{f}\}$: each box defines the pedestrian's position via top-left and bottom-right corner coordinates. (3) Ego-vehicle speed $S=\{s^{f-N+1}, s^{f-N+2},\ldots, s^{f}\}$: represents vehicle's motion. (4) Global semantic map $G_s=\{g_s^{f-N+1}, g_s^{f-N+2},\ldots, g_s^{f}\}$: captures road and traffic context, generated by SegFormer model [10] pretrained on Cityscapes [11]. (5) Global depth map $G_d=\{g_d^{f-N+1}, g_d^{f-N+2},\ldots, g_d^{f}\}$: encodes global distance information, generated using the pretrained Depth Anything v2 model [12]. (6) Local RGB image $L_{RGB}=\{l_{RGB}^{f-N+1}, l_{RGB}^{f-N+2},\ldots, l_{RGB}^{f}\}$ : captures pedestrian appearance, obtained by cropping the original image to the bounding box. (7) Local depth map $L_d=\{l_d^{f-N+1}, l_d^{f-N+2},\ldots, l_d^{f}\}$: captures the pedestrian's relative distance, obtained by cropping the global depth map to the bounding box. Each modality sequence has a length of $N$=16.

## 3 Methodology

The proposed P-MFNet integrates both visual and motion modalities from traffic sce-nes, leveraging multimodal data. The overall architecture is shown in Fig. 1.

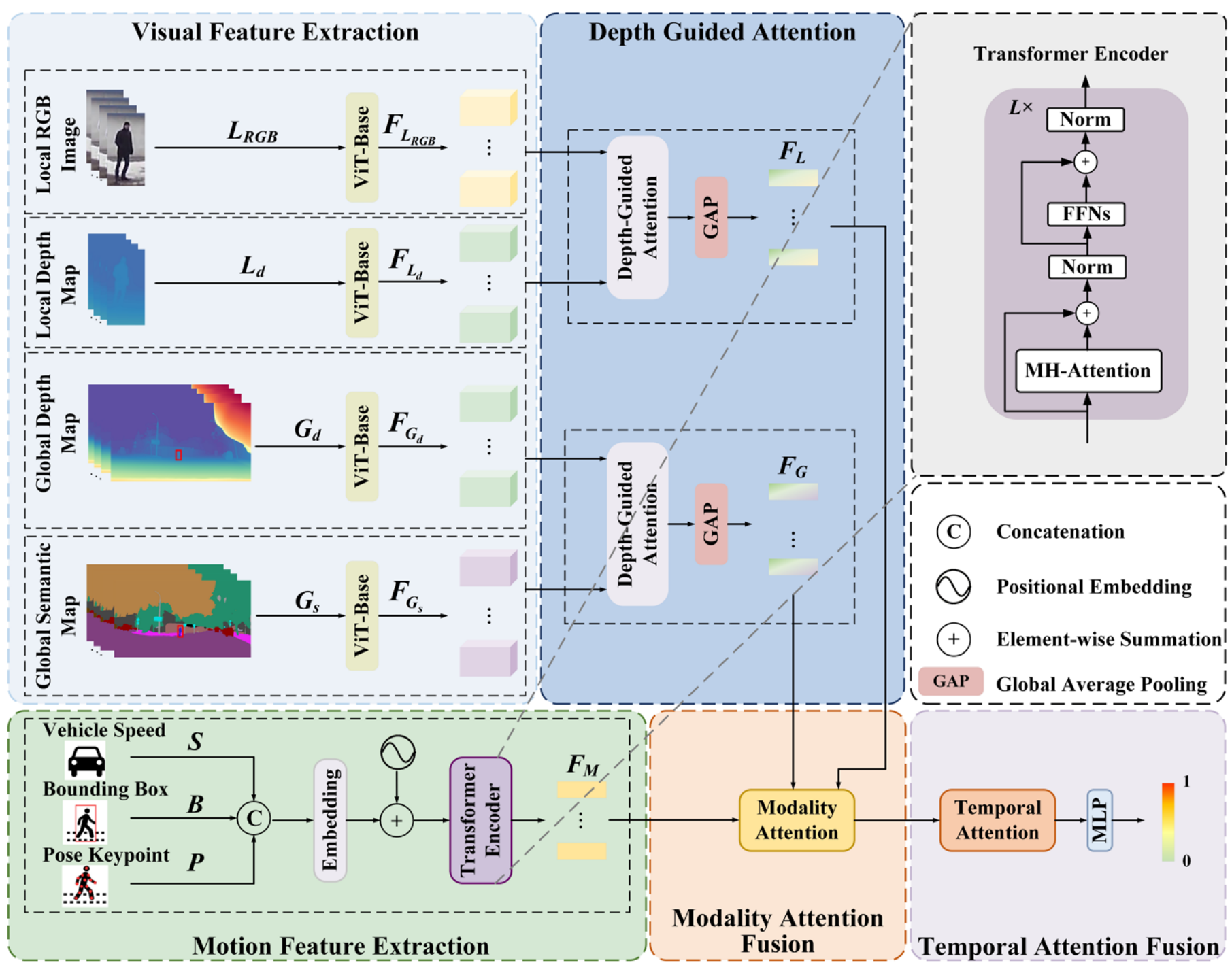

**Fig. 1.** Architecture of the proposed P-MFNet

### 3.1 Visual Feature Extraction

The visual feature extraction (VFE) module extracts the pedestrian-specific cues and global contextual cues from the visual modalities. Inspired by the superior performance of Transformers over CNNs in vision tasks, the Transformer-based backbones are adopted in VFE module. The Vision Transformer (ViT) model [13], pretrained on ImageNet [14], is selected as the backbone. By leveraging self-attention across the entire image, ViT effectively captures fine-grained details and global context, enhancing feature representation. To ensure compatibility with ViT, all the images are resized to 224×224. The final-layer feature maps of each modality's backbone, denoted as $F_{L_{RGB}}$, $F_{L_d}$, $F_{G_s}$, and $F_{G_d}$ are utilized for the visual feature fusion.

### 3.2 Motion Feature Extraction

The motion feature extraction (MFE) module is designed to construct a joint dynamics representation by processing pedestrian's pose keypoints, bounding box, and ego-vehicle speed. The MFE module employs a Transformer encoder [15] to extract motion-related features from sequential pedestrian–vehicle interactions. The three motion modality data from $N$ frames are stacked into an $N$×41 vector, which are then processed by an embedding layer to produce a higher-dimensional representation of size $N$×256. Positional encoding is applied to preserve temporal order, followed by a Transformer encoder to extract motion features. Finally, the motion feature $F_M \in \mathbb{R}^{N\times256}$ from MFE module is fed into the modality attention fusion module.

### 3.3 Depth Guided Attention

The depth-guided attention (DGA) module leverages depth maps to highlight critical pedestrian body parts or salient global entities. DGA module is implemented based on depth-guided attention mechanism. Fig. 2 illustrates its application in the spatial feature fusion of $F_{L_{RGB}}$ and $F_{L_d}$. At time step $\tau$, co-registered spatial feature map $f^{\tau}_{L_{RGB}}$ and $f^{\tau}_{L_d}$ are downsampled by a 1×1 convolution (256 filters) to obtain compact representations $f'^{\tau}_{L_{RGB}}$ and $f'^{\tau}_{L_d}$. These feature maps are then processed as follows:

(1) Depth-guided channel attention (DGCA): DGCA enhances important RGB feature channels using channel attention maps derived from the depth feature map $f'^{\tau}_{L_d}$, which is processed via spatial-wise global average and max pooling, followed by a sigmoid-activated fully connected layer:

$$CA^{\tau} = \sigma\left(FC\left(\mathrm{AvgPool}_s(f'^{\tau}_{L_d}) + \mathrm{MaxPool}_s(f'^{\tau}_{L_d})\right)\right) \quad (1)$$

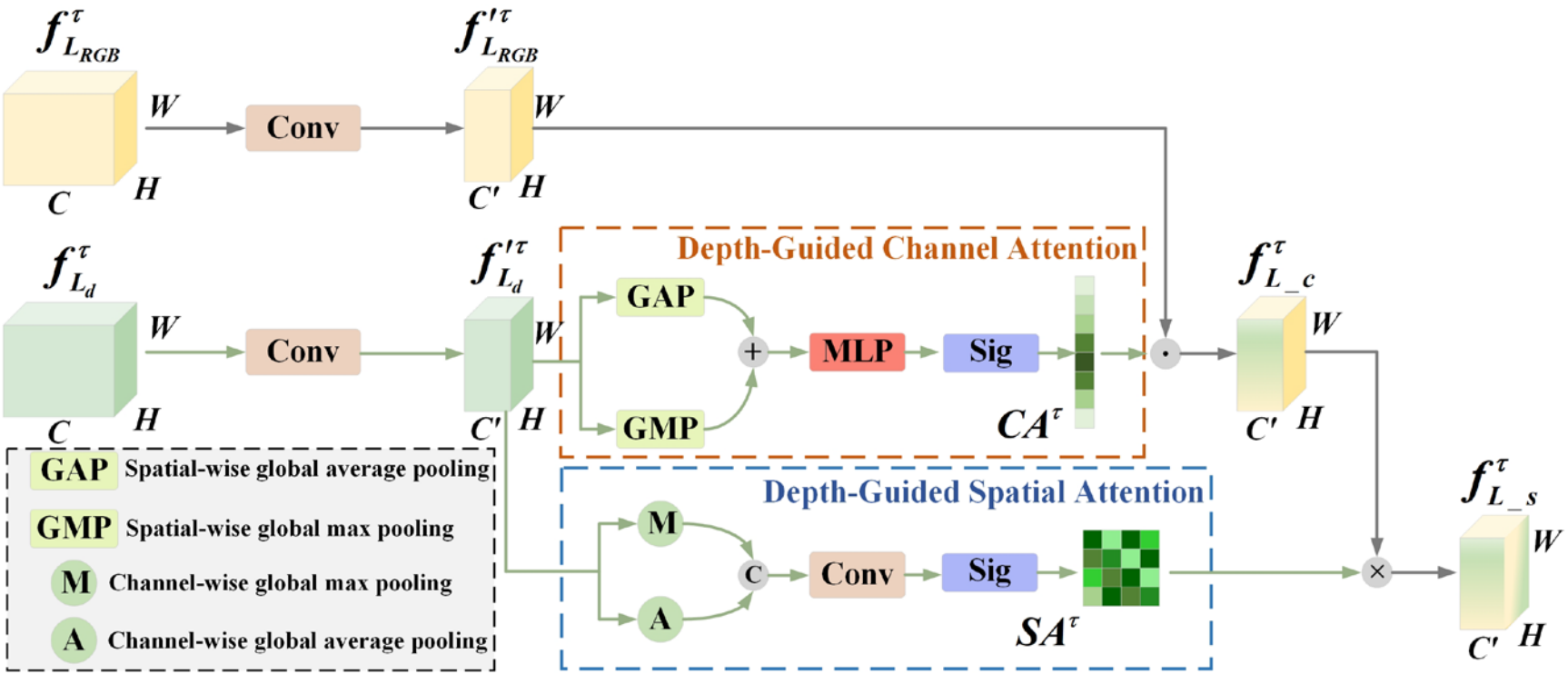

**Fig. 2.** Architecture of the depth-guided attention

The channel-wise attention map $CA^{\tau}$ is then multiplied with the $f'^{\tau}_{L_{RGB}}$ to obtain the channel-enhanced feature map $f^{\tau}_{L_c}$.

(2) Depth-guided spatial attention (DGSA): DGSA enhances spatially relevant regions using spatial attention maps derived from the depth feature map $f'^{\tau}_{L_d}$. Channel-wise global average and max pooling are applied, concatenated, and processed by a 3×3 convolution with sigmoid activation to produce spatial attention maps:

$$SA^{\tau} = \sigma\left(\mathrm{Conv}\left(\left[\mathrm{AvgPool}_c(f'^{\tau}_{L_d}); \mathrm{MaxPool}_c(f'^{\tau}_{L_d})\right]\right)\right) \quad (2)$$

The spatial attention map $SA^{\tau}$ is multiplied with $f^{\tau}_{L_c}$ to obtain the spatial-enhanced feature map $f^{\tau}_{L_s}$. $F_{G_s}$, and $F_{G_d}$ are processed in a similar manner. The enhanced featur-e maps are then pooled for modality attention fusion.

### 3.4 Modality Attention Fusion

The modality attention fusion (MAF) module integrates multiple modality features by dynamically weighting their importance at each time step. Let $f_m^\tau \in \mathbb{R}^d$ denotes the feature of modality $m \in \{M, L, G\}$. The attention score for modality $m$ is calculated as:

$$e_m^\tau = W_s^T \tanh(W_e f_m^\tau + b_e) + b_s \tag{3}$$

Where $W_e \in \mathbb{R}^{d\times d}$ and $W_s \in \mathbb{R}^d$ are learnable parameters, $b_e \in \mathbb{R}^d$ and $b_s \in \mathbb{R}$ are their corresponding bias terms.

The scores are normalized via the softmax function to obtain attention weights:

$$\alpha_m^\tau = \frac{\exp(e_m^\tau)}{\sum_{n=1}^{3} \exp(e_n^\tau)} \tag{4}$$

Finally, the fused feature at each time step is computed as a weighted sum of $f_M^\tau$, $f_L^\tau$ and $f_G^\tau$. The output is then passed through a fully connected layer followed by a tanh activation function to produce the final fused representation:

$$f^\tau = \tanh\left( W_c \sum_{m=1}^{3} \alpha_m f_m^\tau \right) \tag{5}$$

### 3.5 Temporal Attention Fusion

The temporal attention fusion (TAF) module leverages a Transformer encoder [15] to model temporal dependencies among sequential features, using multi-head self-attention (MHA) and feed-forward network (FFN).

The MHA sub-layer comprises $m$ parallel attention heads, allowing it to simultaneously attend to different representation subspaces. The inputs of attention head $i$ are query $Q_i = QW_i^Q$, key $K_i = KW_i^K$ and value $V_i = VW_i^V$, where $W_i^Q$, $W_i^K$ and $W_i^V$ are trainable weight parameters. It then computes the attention matrix using scaled dot products between the queries and keys, capturing the correlations within the feature sequence. By multiplying $V_i$ with the attention matrix, each output feature incorporates temporal information from other parts of the sequence:

$$SA_i(Q_i, K_i, V_i) = \text{Softmax}\left( \frac{Q_i K_i^T}{\sqrt{d_k}} \right) V_i \tag{6}$$

All self-attention heads are concatenated and multiplied by a trainable parameter $W^O \in \mathbb{R}^{md_v \times D}$ to generate the output of MHA sub-layer:

$$MHA(Q, K, V) = [SA_1, SA_2, ..., SA_m] W^O \tag{7}$$

The feature of the last frame from the final layer of the TAF module is then passed through an MLP to predict the pedestrian's crossing intention.

# 4 Experiments

## 4.1 Datasets and Implementation Details

P-MFNet is trained and evaluated on the JAAD dataset, which includes front-view vehicle videos recorded in natural conditions. The JAAD dataset is split into two subsets: $JAAD_{beh}$, consisting of 495 crossing pedestrians and 191 non-crossing pedestrians, and $JAAD_{all}$, which includes an additional 2,100 visible pedestrians without crossing intention. We follow the same data split and overlap ratio as the benchmark [2]. Performance is evaluated using five widely adopted metrics: accuracy, precision (P), recall (R), F1 score, and AUC.

P-MFNet takes a sequence of 16 frames, i.e., $N$=16. We applied a dropout rate of 0.2, L2 regularization of 0.001 in the MLP layer, binary cross-entropy loss, and the Adam optimizer with a learning rate of $2\times10^{-5}$. The Transformer encoder is configured with 2 layers and 4 attention heads in both the MFE and TAF modules.

## 4.2 Quantitative Results

To evaluate the effectiveness of P-MFNet, we compare it with several baseline methods, as shown in Table 1. The best and second-best results are highlighted in green and blue, respectively. On the $JAAD_{beh}$ dataset, P-MFNet achieves the highest accuracy, AUC, and F1 score. Specifically, it outperforms the second-best method by 1% in accuracy (compared to STMA-GCN [21]), 3% in AUC (compared to II-GRU [22] and PPCIatt [4]), and matches the F1 score of STMA-GCN [21]. In terms of precision and recall, P-MFNet ranks second, falling behind V-PedCross [6] by 1% in precision and behind STMA-GCN [21] by 4% in recall.

On the $JAAD_{all}$ dataset, P-MFNet attains the highest accuracy of 89% and a precision of 70%, which are 2% and 10% higher, respectively, than those of the second-best method, II-GRU [22]. P-MFNet ranks second in F1 score, falling 9% behind PP-

**Table 1.** Performance comparison with baseline methods on JAAD dataset.

| Models | $JAAD_{beh}$ | | | | | $JAAD_{all}$ | | | | |
|---|---|---|---|---|---|---|---|---|---|---|
| | **Acc** | **AUC** | **F1** | **P** | **R** | **Acc** | **AUC** | **F1** | **P** | **R** |
| SF-GRU[7] | 0.58 | 0.56 | 0.65 | 0.68 | 0.62 | 0.76 | 0.77 | 0.53 | 0.40 | 0.79 |
| FUSSI[16] | 0.59 | 0.58 | 0.69 | 0.66 | 0.73 | 0.60 | 0.72 | 0.40 | 0.27 | 0.73 |
| SingleRNN [17] | 0.60 | 0.54 | 0.70 | 0.65 | 0.76 | 0.78 | 0.77 | 0.54 | 0.42 | 0.75 |
| BiPed [18] | - | - | - | - | - | 0.83 | 0.79 | 0.60 | 0.52 | - |
| IntFormer [19] | 0.59 | 0.54 | 0.69 | - | - | 0.86 | 0.78 | 0.62 | - | - |
| PCPA [2] | 0.56 | 0.54 | 0.63 | 0.66 | 0.60 | 0.77 | 0.79 | 0.56 | 0.42 | **0.83** |
| Global PCPA [5] | 0.62 | 0.54 | 0.74 | 0.65 | 0.85 | 0.83 | **0.82** | 0.63 | 0.51 | **0.81** |
| MMH-PAP [20] | - | - | - | - | - | 0.84 | 0.80 | 0.62 | 0.54 | - |
| V-PedCross [6] | 0.61 | 0.50 | 0.75 | **0.71** | 0.80 | 0.82 | 0.74 | 0.64 | 0.58 | 0.63 |
| STMA-GCN[21] | **0.69** | 0.58 | **0.80** | 0.68 | **0.97** | - | - | - | - | - |
| II-GRU [22] | 0.68 | **0.60** | 0.78 | 0.68 | 0.91 | **0.87** | **0.81** | 0.65 | **0.60** | 0.71 |
| PPCIatt [4] | 0.67 | **0.60** | 0.77 | - | - | 0.81 | 0.78 | **0.75** | - | - |
| P-MFNet | **0.70** | **0.63** | **0.80** | **0.70** | **0.93** | **0.89** | 0.79 | **0.66** | **0.70** | 0.63 |

CIatt [4]. In terms of AUC, P-MFNet ranks fourth, falling 3% behind the best method, Global PCPA [5]. The quantitative results indicate that the proposed P-MFNet network benefits from rich cross-modal interactions and efficient fusion strategies, facilitating the extraction and integration of complementary features and ultimately leading to superior performance over baseline methods.

### 4.3 Ablation Study

Ablation studies are conducted to investigate the contributions of different modules and to compare various multimodal fusion strategies. The results of the ablation study are summarized in Table 2.

To assess the effectiveness of different backbones, P-MFNet-V1 replaces the ViT-Base with VGG16 [1]. P-MFNet-V1 demonstrates a consistent performance drop across all evaluation metrics, with accuracy decreasing by 4% and 5% on $JAAD_{beh}$ and $JAAD_{all}$ datasets, respectively. ViT-Base demonstrates superior effectiveness in capturing both pedestrian-relevant features and global scene context.

To evaluate the effectiveness of different motion feature extraction methods, P-MFNet-V2 replaces the Transformer with gated recurrent unit (GRU). P-MFNet-V2 shows a slight decrease in accuracy, dropping by 1% and 3% on $JAAD_{beh}$ and $JAAD_{all}$ datasets, respectively. The performance gap indicates that the Transformer extracts motion features more effectively than GRU.

To assess the contribution of motion modalities and visual modalities, P-MFNet-V3 uses only visual modalities, while P-MFNet-V4 uses only motion modalities as input. Both P-MFNet-V3 and P-MFNet-V4 exhibit notable performance drops. The decline is most pronounced in P-MFNet-V4, with accuracy decreasing by 7% and 8% on $JAAD_{beh}$ and $JAAD_{all}$ datasets, respectively, while P-MFNet-V3 shows a smaller drop of 4% and 4%, respectively. These results suggest that visual and motion modalities provide complementary information, and their integration improves pedestrian intention prediction, with visual cues contributing more significantly.

To examine the impact of depth map and DGA module, P-MFNet-V5 removes the depth map, while P-MFNet-V6 replaces the depth-guided attention with simple feature addition. P-MFNet-V5 exhibits a significant performance drop due to the absence of depth maps, which deprives the model of crucial spatial cues. In P-MFNet-V6, all metrics except recall decline, suggesting that simple feature addition fails to exploit spatial relationships, limiting the fusion of complementary visual information.

To evaluate the impact of MAF module, P-MFNet-V7 replaces modality attention with simple addition. We observe a performance drop of 2% in accuracy on both datasets. It is due to the lack of modality attention, which treats all modalities equally and fails to account for their varying importance.

**Table 2.** Ablation study results on the JAAD dataset.

| Models | $JAAD_{beh}$ | | | | | $JAAD_{all}$ | | | | |
|---|---|---|---|---|---|---|---|---|---|---|
| | **Acc** | **AUC** | **F1** | **P** | **R** | **Acc** | **AUC** | **F1** | **P** | **R** |
| P-MFNet-V1 | 0.66 | 0.58 | 0.76 | 0.67 | 0.88 | 0.84 | 0.75 | 0.57 | 0.53 | 0.61 |
| P-MFNet-V2 | **0.69** | 0.61 | **0.79** | 0.69 | 0.92 | 0.86 | 0.78 | 0.62 | 0.58 | 0.67 |
| P-MFNet-V3 | 0.66 | 0.57 | 0.77 | 0.67 | 0.92 | 0.85 | 0.77 | 0.60 | 0.56 | 0.64 |
| P-MFNet-V4 | 0.63 | 0.50 | 0.77 | 0.63 | **1.00** | 0.81 | **0.80** | 0.59 | 0.47 | **0.78** |
| P-MFNet-V5 | 0.68 | 0.60 | **0.79** | 0.68 | 0.94 | 0.84 | **0.80** | 0.62 | 0.54 | **0.72** |
| P-MFNet-V6 | 0.68 | 0.58 | **0.79** | 0.67 | **0.97** | 0.86 | 0.78 | 0.63 | 0.59 | 0.67 |
| P-MFNet-V7 | 0.68 | **0.62** | 0.77 | **0.70** | 0.87 | **0.87** | **0.80** | **0.66** | **0.62** | 0.69 |
| P-MFNet | **0.70** | **0.63** | **0.80** | **0.70** | 0.93 | **0.89** | 0.79 | **0.66** | **0.70** | 0.63 |

### 4.4 Qualitative Results

Fig. 3 presents several qualitative results, where P-MFNet is primarily compared with PCPA and Global PCPA. The target pedestrian is indicated by a red bounding box. “GT” stands for ground truth, with “C” indicating crossing and “NC” indicating no crossing. The results demonstrate that P-MFNet consistently achieves accurate predictions of pedestrian crossing intentions, whereas PCPA and Global PCPA encounter some difficulties. The competitive performance of P-MFNet stems from its effective cross-modal interactions and multimodal fusion strategy, which jointly facilitate the extraction of subtle cues, leading to more accurate pedestrian intention prediction.

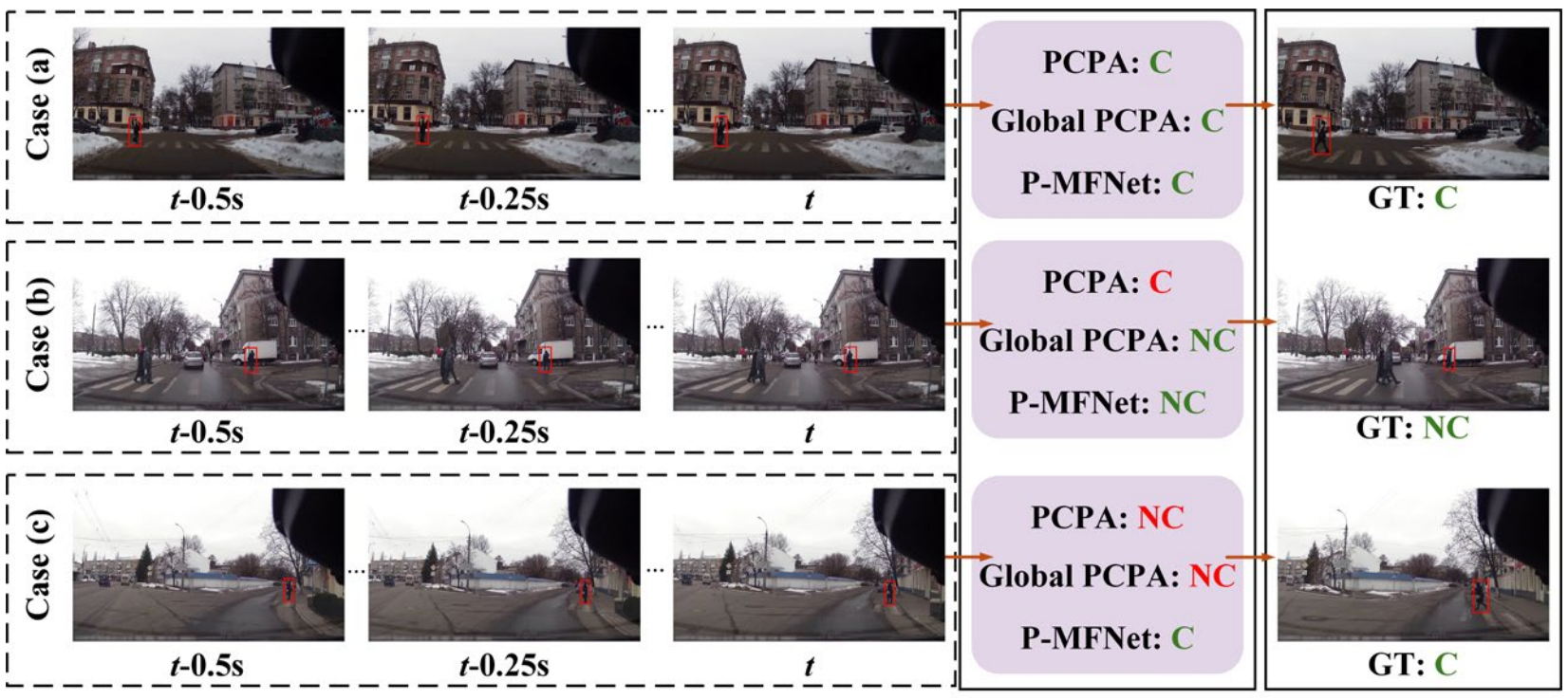

**Fig. 3.** Qualitative results of PCPA, Global PCPA and P-MFNet on JAAD dataset

Fig. 4 illustrates the visualization of temporal attention maps across consecutive frames on $JAAD_{beh}$ and $JAAD_{all}$ datasets, highlighting the relevance of each frame in temporal feature fusion. It shows that different layers and attention heads extract diverse features, attending to different time steps within the temporal sequence. More a-ttention is observed on frames 12 to 16, which corresponds to the fact that the final frames provide richer cues of pedestrian crossing intention.

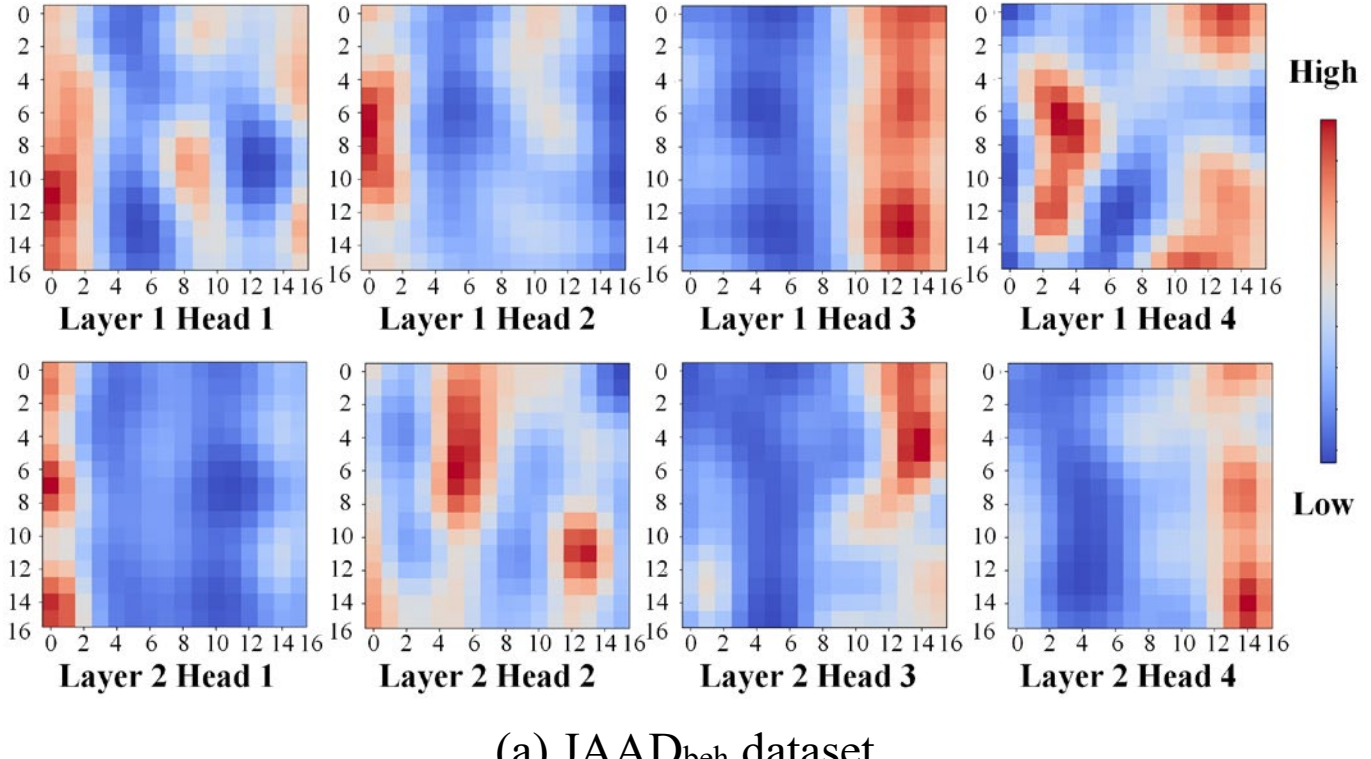

(a) $JAAD_{beh}$ dataset

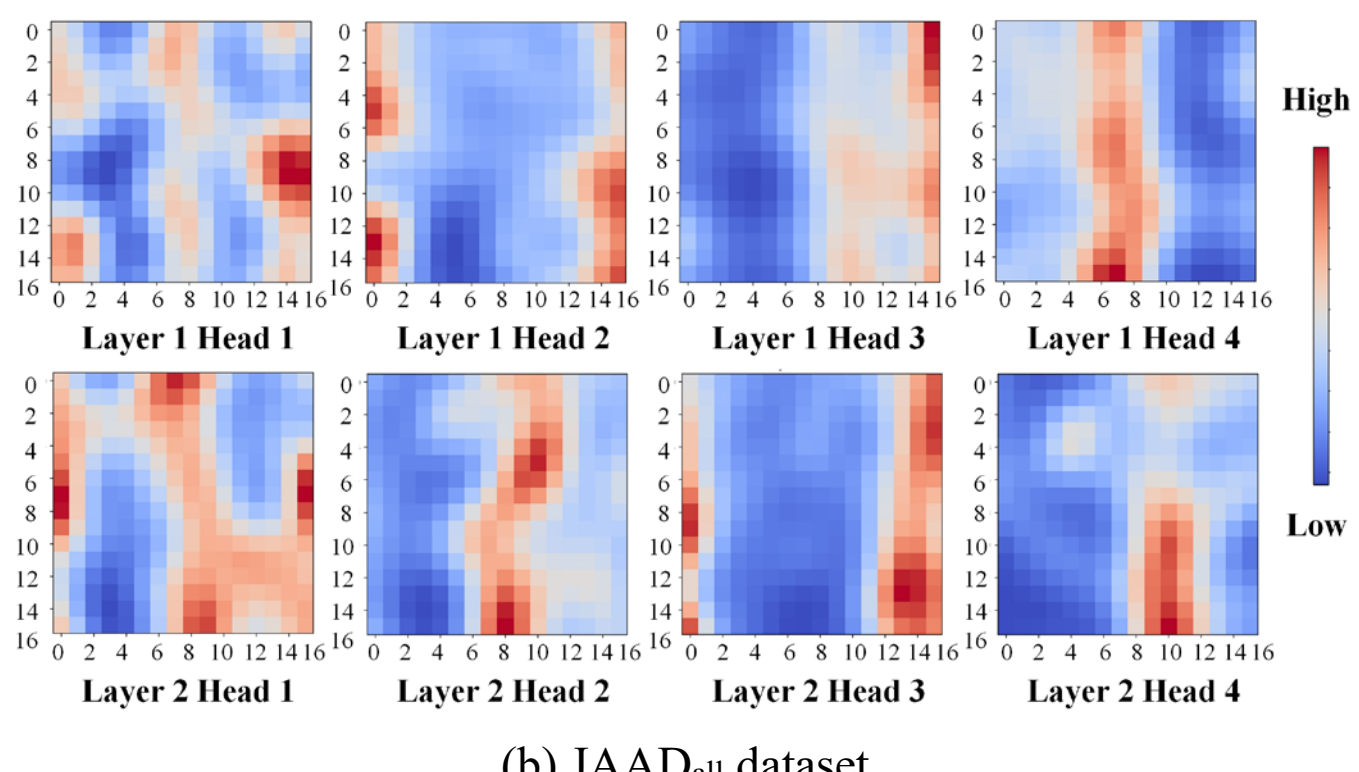

(b) $\text{JAAD}_{\text{all}}$ dataset

**Fig. 4.** Visualization of temporal attention maps across consecutive frames

## 5 Conclusions

In this paper, we propose a multimodal fusion network for pedestrian crossing intention prediction. Transformer-based visual and motion feature extraction modules are designed to extract complementary cues from different modalities. A depth-guided attention module is designed to leverage depth maps to guide attention toward specific body parts and the global context. A modality attention module and a temporal attention module are introduced to perform modality fusion and temporal feature integration. The proposed network achieves 70% and 89% overall accuracy on $\text{JAAD}_{\text{beh}}$ and $\text{JAAD}_{\text{all}}$ datasets, respectively, which surpasses all 12 baseline methods. Ablation studies are performed to validate the effectiveness and design rationale of the proposed network architecture. Future work will focus on model lightweighting and real-world deployment.